\documentclass{article}
\usepackage{spconf,amsmath,graphicx}
\usepackage{booktabs}
\usepackage{amsfonts}

\title{Nana-HDR: A Non-attentive Non-autoregressive Hybrid Model for TTS}
\name{Shilun Lin, Wenchao Su, Li Meng, Fenglong Xie, Xinhui Li, Li Lu}
\address{Tencent, Beijing, China}
\begin{document}
%
\maketitle
\begin{abstract}
This paper presents Nana-HDR, a new \textbf{n}on-\textbf{a}ttentive \textbf{n}on-\textbf{a}utoregressive model with \textbf{h}ybrid Transformer-based \textbf{D}ense-fuse encoder and \textbf{R}NN-based decoder for TTS. It mainly consists of three parts: Firstly, a novel Dense-fuse encoder with dense connections between basic Transformer blocks for coarse feature fusion and a multi-head attention layer for fine feature fusion. Secondly, a single-layer non-autoregressive RNN-based decoder. Thirdly, a duration predictor instead of an attention model that connects the above hybrid encoder and decoder. Experiments indicate that Nana-HDR gives full play to the advantages of each component, such as strong text encoding ability of Transformer-based encoder, stateful decoding without being bothered by exposure bias and local information preference, and stable alignment provided by duration predictor. Due to these advantages, Nana-HDR achieves competitive performance in naturalness and robustness on two Mandarin corpora.
\end{abstract}
\section{Introduction}
\label{sec:intro}
With advances in deep learning, the text-to-speech acoustic model that converts text into acoustic features gradually transits from Hidden Markov Models (HMMs) to Deep Neural Network (DNN) based ones. Sequence-to-sequence neural network with attention mechanism \cite{sutskever2014sequence} is one of the most popular methods. The traditional synthesis process is simplified by merging the generation of linguistic feature and acoustic feature into a single network. Works such as Char2Wav \cite{sotelo2017char2wav}, Deep Voice 3 \cite{ping2018deep} and Tacotron \cite{wang2017tacotron}, make significant progress in generating highly natural speech close to human quality.

As an attention-based autoregressive model, Tacotron is able to generate human-like speech for in-domain text. However, it cannot handle out-domain situations robustly. For instance, the length of testing text is quite different from that of training text and the field of testing text is not included in the training set. The causes of the robustness issue can be roughly classified as follows: Firstly, there are no explicit restrictions on the soft attention mechanisms such as content-based attention \cite{bahdanau2015neural} and location sensitive attention \cite{shen2018natural} to prevent skip, repetition and mispronunciation. Secondly, the model predicts a stop flag to judge whether the synthesis process is completed. Therefore, a wrong prediction can lead to serious failures, such as early cut-off and late stop. Finally, teacher forcing training induces a mismatch between training and inference, usually known as exposure bias \cite{2016SEQUENCE}. And the local information preference on autoregressive decoder may weaken the dependence between predicted acoustic features and text conditions, which makes the model tend to produce bad cases \cite{liu2019maximizing}. Local information preference still exists due to autoregressive property, although teacher forcing is not applied \cite{chen2016variational}. 

Many efforts are made to solve the above problems. Methods in \cite{graves2013generating, zhang2018forward, battenberg2020location} improve the attention mechanism by introducing location information and monotonic constraints. These approaches are proved to be effective in improving the speed of convergence, the stability of feature generation and the robustness of long sentence synthesis. However, the improved attention mechanisms can not fundamentally solve the problem of attention failure. Fastspeech \cite{ren2019fastspeech} relies on the duration predictor instead of attention model, which eliminates the robustness problems caused by attention failure and stop frame misprediction. As a feed-forward non-autoregressive model, Fastspeech can instantly convert text into acoustic features. However, there is still a gap between its synthetic quality and that of autoregressive models. For exposure bias and local information preference issues, the effect of exposure bias on the autoregressive decoder is reduced by adversarial training \cite{guo2019new}. Inspired by InfoGAN \cite{chen2016infogan}, a recognizer is introduced to maximize the mutual information between predicted feature and text condition which reduces the impact of local information preference \cite{liu2019maximizing}.

Tacotron and Fastspeech mentioned above have single type of encoder and decoder. Experiments demonstrate that, compared to the architecture with a single type of encoder and decoder, the hybrid architecture with Transformer-based encoder and RNN-based decoder achieves the best results on the English-to-French machine translation task \cite{chen2018best}. The results confirm their intuition that Transformer-based encoder is good at text feature extraction and the stateful RNN-based decoder is beneficial for conditional generation. 

\begin{figure*}[ht]
  \centering
  \includegraphics[scale=0.48]{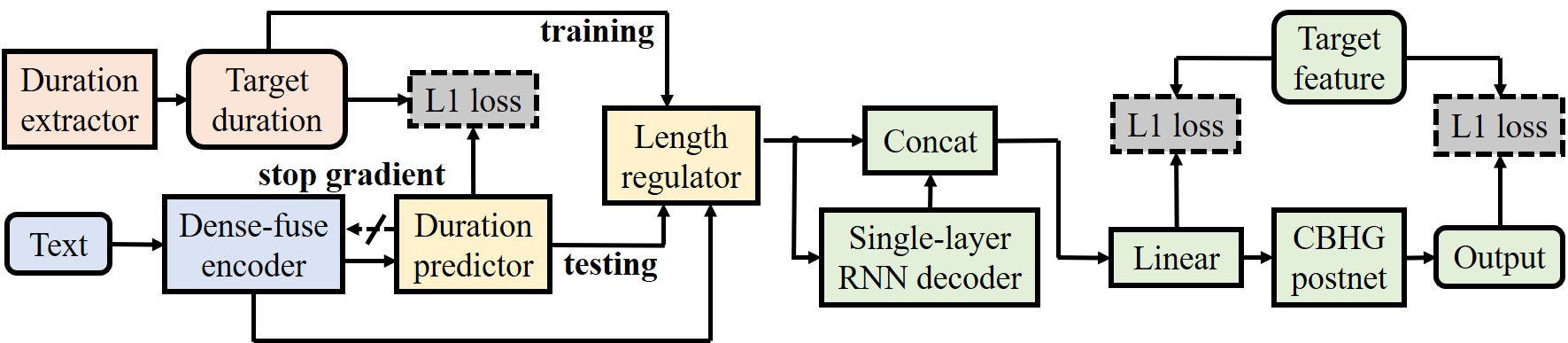}
  \caption{Architecture of the proposed Nana-HDR.}
  \label{fig:Nana-HDR}
\end{figure*}
Recently, the work in the field of natural language processing \cite{jawahar2019does} further explores what BERT learn about the structure of language. The results indicate that intermediate layers of BERT encode a rich hierarchy of linguistic information, with surface features at the bottom, syntactic features in the middle and semantic features at the top. This conclusion is similar to the one obtained by visual analysis of Convolutional Neural Network (CNN), which shows that the features learned by CNN are also hierarchical \cite{zeiler2014visualizing}. The bottom layers learn the surface features like edges, corners and colors. Higher layers learn more semantic information such as faces. In the field of computer vision, the performance of various tasks can be improved by fusing features with different representation meanings. ResNet \cite{he2016deep} and DenseNet \cite{huang2017densely} merge the features of different layers through bypass connections, which improves the performance of image classification while alleviating gradient vanishing. FCN \cite{long2015fully}, U-Net \cite{ronneberger2015u} and SegNet \cite{badrinarayanan2017segnet} fuse the features of different encoder layers in different ways, and then provide the fusion feature to the decoder, which effectively improve the accuracy of semantic segmentation. Because the features extracted by Transformer encoder and CNN have similar hierarchical characteristics, the feature fusion method for CNN may be transferred to Transformer encoder for the purpose of improving the performance of the TTS acoustic model.

In this paper, a non-attentive non-autoregressive model with hybrid Transformer-based Dense-fuse encoder and RNN-based decoder (Nana-HDR) for TTS is proposed. The main contributions include three parts:

1) We propose a novel Dense-fuse encoder based on the Transformer encoder. The dense-fuse here has two meanings. Firstly, similar to DenseNet which densely connects multiple convolution modules through bypass connections, transformer blocks are connected through dense bypass-connections, which enables the bottom layer features to be reused by the top layer. Secondly, the features learned by different transformer blocks are fused through a learnable multi-head attention layer on the top. The features with different representation meanings are fused twice (coarse and fine fusion) which makes the training of the duration extractor more stable.  Meanwhile, the Dense-fuse encoder can extract text encoding with strong representation capability which is beneficial to improve naturalness and robustness.

2) We verify that it is feasible to use a non-autoregressive single-layer RNN as decoder. Here non-autoregressive means that the decoder does not rely on the autoregressive feedback of predicted or real acoustic features. In this way, the influence of the exposure bias and the local information preference can be eliminated without introducing auxiliary strategies like scheduled sampling \cite{bengio2015scheduled} and maximum mutual information. Further, the efficiency of decoding is significantly improved.

3) The duration predictor is used to connect the Dense-fuse encoder and the non-autoregressive RNN decoder, which avoids the instability caused by the attention model. In this paper, non-attentive means that the encoder and decoder of Nana-HDR are not connected through the attention model. Such a non-attentive non-autoregressive hybrid model gives full play to the advantages of various components. In addition to achieving good in-domain naturalness, the model can deal with out-domain scenarios robustly.

\section{Proposed Method}
\label{sec:pm}
Figure~\ref{fig:Nana-HDR} shows our proposed Nana-HDR which consists of three major components: (1) A novel Dense-fuse encoder, which densely connects multiple basic Transformer blocks through bypass connections. The outputs of each Transformer block with different representation meaning are fused in a learnable way. (2) A one-layer RNN-based decoder without explicit autoregression which speeds up inference while alleviating exposure bias and local information preference. (3) A duration predictor  (with a length regulator) instead of attention model that connects two hybrid parts. Collapse caused by attention failure and stop flag misprediction can be avoided by this non-attentive structure.
\begin{figure}[!htb]
  \centering
  \includegraphics[scale=0.44]{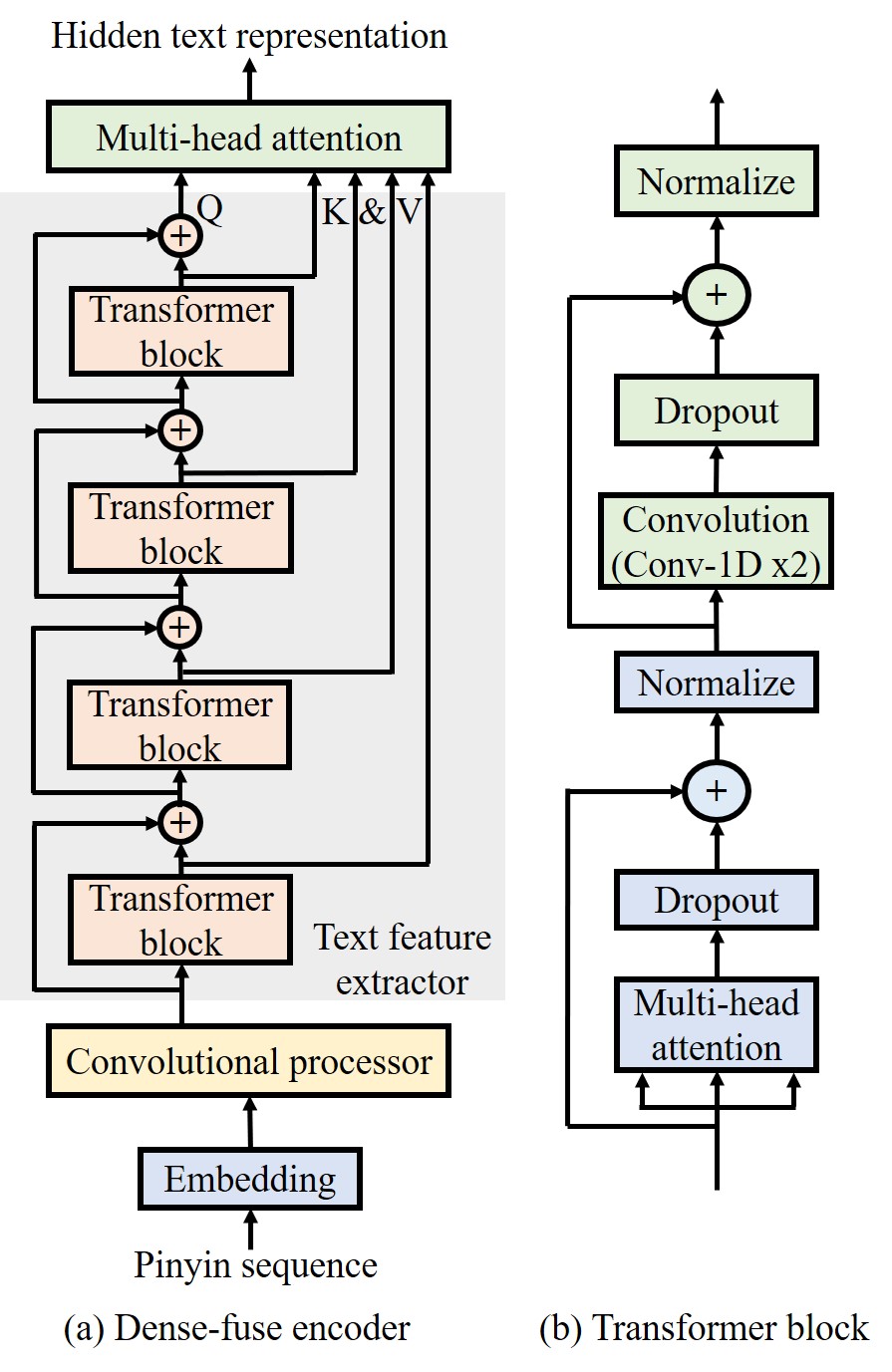}
  \caption{Architecture of the proposed Dense-fuse encoder.}
  \label{fig:densefuse}
\end{figure}

\subsection{Dense-fuse encoder}
The proposed Dense-fuse encoder is presented in Figure~\ref{fig:densefuse}(a). Chinese Pinyin sequence with tone and prosody information is converted to the corresponding embedding sequence. Then, a convolutional processor with multiple Conv-1D layers are applied to preprocess the embedding sequence. For two reasons, sinusoidal positional embedding is not used. Firstly, the maximum length of the input sequence needs to be preset for calculating of the sinusoidal encoding table, which makes the model unable to handle arbitrarily long sentences. Secondly, using convolutional layers as input processor is able to capture implicit relative position information which has been proved to improve the performance of automatic speech recognition \cite{mohamed2019transformers}. 

Next, the output of the processor with convolutional context is fed to the text feature extractor stacked by basic Transformer blocks. As shown in Figure~\ref{fig:densefuse}(b), each Transformer block is composed of multi-head attention and convolution sub-modules. All sub-modules follow a strict computation flow: process, dropout, residual-add and layer normalize. Transformer blocks are densely connected through bypass connections. From the perspective of back propagation, the supervision signal from the top layer can be better transmitted back to the lower layer, which plays a similar role of deep supervision and makes the model easier to train. From the perspective of feature fusion, dense connections make the features of lower layer can be reused by higher layer. Since the features extracted by different layers of Transformer encoder have different representation meanings \cite{jawahar2019does}, intuitively such feature reuse can enhance the representation ability of the final encoding as it has been proven in DenseNet. Different from DenseNet which uses channel-wise concatenation, element-wise addition is applied for feature fusion which helps to maintain the computational complexity by keeping encoding dimension between blocks unchanged. In this way, features of different layers are treated indiscriminately which can be regarded as a coarse fusion. It is relatively rough to fuse the features learned by transformer blocks through dense bypass-connections, because the model cannot control the proportion of these features during the fusion. 

In \cite{wang2018style}, attention mechanism is applied to output a set of combination weights over the style tokens. Inspired by this method, multi-head attention is introduced for fine fusion. As shown in Figure~\ref{fig:densefuse}(a), the coarse fusion feature is used as the query (Q), and the output of each Transformer block is used as the key (K) and value (V) of an attention head. The final hidden text representation is obtained by combining outputs of different blocks with the weights learned by multi-head attention layer. Fine fusion can be regarded as a further adjustment of the coarse fusion feature by the model. The model can decide how much additional information the coarse fusion feature needs to obtain from each Transformer block by fine fusion.
\begin{figure}[!htb]
  \centering
  \setlength{\abovecaptionskip}{0.2cm}
  \includegraphics[scale=0.45]{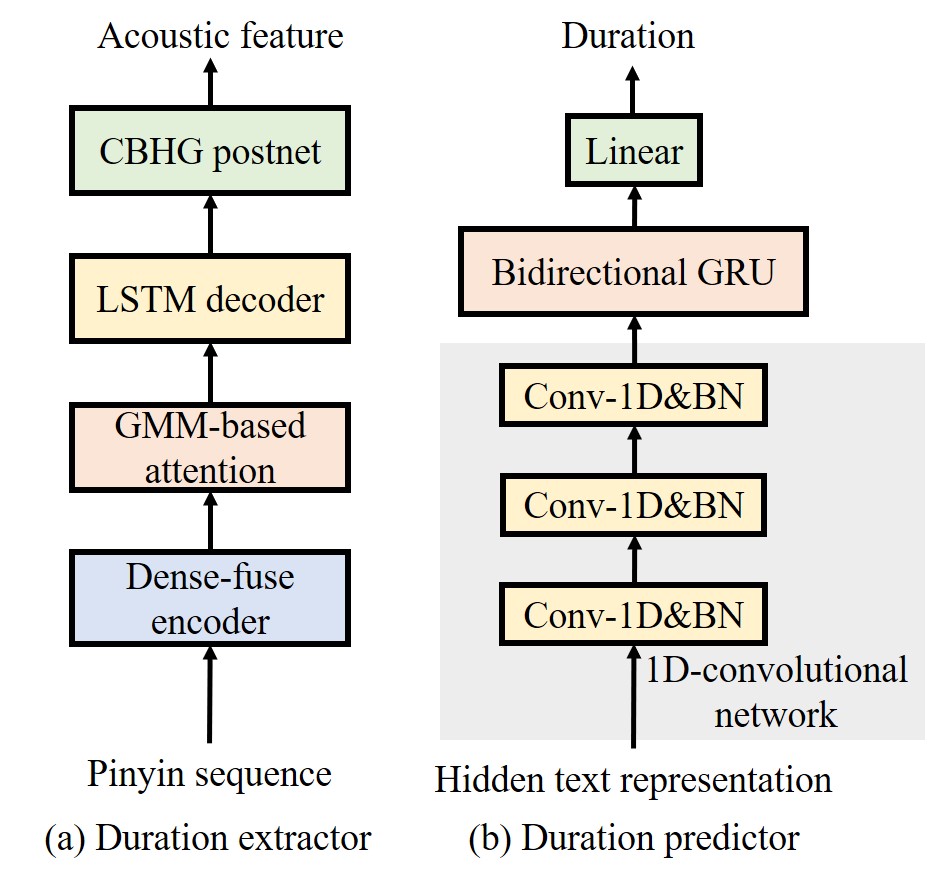}
  \caption{Architecture of the duration extractor and predictor.}
  \label{fig:duration}
\end{figure}

\subsection{Duration extractor and predictor}
Before training Nana-HDR, an attention-based duration extractor is trained to align the training text and training audio, and then output the duration of the training text tokens (target duration). Duration of individual text token can be obtained by ASR force alignment. However, our off-the-shelf Mandarin speech recognition systems are trained with phoneme-level text token. It is time-consuming to train an efficient speech recognition model (based on Pinyin character-level text token) from scratch. Training an attention-based model and extracting target duration through ground truth aligned (GTA) mode is a relatively efficient and accurate way. As shown in Figure~\ref{fig:duration}(a), the duration extractor is a sequence-to-sequence model composed of the above-mentioned Dense-fuse encoder, GMM-based attention model, a two-layers LSTM decoder and a CBHG postnet. As shown in Figure~\ref{fig:Nana-HDR}, the extracted target duration is used to upsample the encoder outputs (by the length regulator) to match the length of target acoustic features during training Nana-HDR. It is also used as label for the learning of the duration predictor. 

The duration predictor is trained jointly as a part of Nana-HDR. As shown in Figure~\ref{fig:duration}(b), the duration predictor processes the hidden text representation (output of Nana-HDR's Dense-fuse encoder) through a 1D-convolutional network to capture local position-related information. Considering that the duration is related to the global context information, a bidirectional RNN layer is added to the convolution network, and its output is linearly mapped to a scalar. During training, the gradient of duration predictor is prevented from back-propagating to the Dense-fuse encoder for the purpose of avoiding affecting its learning. During inference, duration predictor combined with length regulator are used to solve the problem of length mismatch between hidden text representation and acoustic feature to be generated.
\subsection{Non-autoregressive single-layer RNN decoder}
Decoder which keeps tracking the state information is beneficial for conditional generation \cite{chen2018best}. As a stateful decoder, autoregressive RNN-based decoder is widely used in text-to-feature model. However, it suffers from exposure bias, local information preference and slow inference. 

Scheduled sampling deals with the exposure bias by adjusting the drop teacher forcing frame rate empirically. For local information preference, it is mainly caused by the powerful RNN decoder with autoregressive dependency. Let $z$ and $x$ be conditional variable obtained from text and corresponding acoustic features to be decoded, the autoregressive conditional decoding process can be described as $ p(x|z) = {\textstyle \prod_{i}^{N}p(x_{i}|z,x_{<i})} $, where N is the number of acoustic frames in $x$. Since RNNs are universal function approximators and any joint distribution over $x$ admits an autoregressive factorization, the RNN autoregressive decoding distribution can in theory represent any probability distribution. The information that can be modeled locally by decoding distribution $ p(x|z) $ will be encoded locally without using information from $z$ and only the remainder will be modeled using them \cite{chen2016variational}. Experiments in \cite{bowman2016generating} show that weakening the autoregressive part of the model by dropout can encourage the conditional variables to be used. Using larger reduction factor (e.g. r=5) in Tacotron works in a similar way, which can alleviate local information preference to a certain extent. However, this method will introduce a problem: how to select the context conditional frame when using the duration-based model. 

\cite{liu2019maximizing} further formalizes conditional autoregressive attention-based model Tacotron as a variational encoder-decoder. The goal of training is to maximize the sum of conditional likelihood of text $t$ and acoustic features $x$ pairs in the training set. For each pair, the conditional likelihood can be written as $\log_{}{p_{\theta}(x|t)}={\textstyle\sum_{i=1}^{N}}\log_{}{p_{\theta} } (x_{i}|x_{<i},t)$. The distribution of the time-aligned latent variables (context vectors generated by attention model) $c$ can be factorized as $\log_{}{p_{\theta } (c|x,t)}= {\textstyle\sum_{i=1}^{N}}\log_{}{p_{\theta} } (c_{i}|x_{<i},t )$ \cite{shankar2018posterior}. Then the conditional likelihood of each training pair can be written as $\log_{}{p_{\theta}(x|t)}=\log_{}{\int_{c}^{}}{p_{\theta} } (x,c|t )dc={\textstyle\sum_{i=1}^{N}}\log_{}{\int_{c_{i}}^{}}{p_{\theta} }(x_{i},c_{i}|x_{<i},t )$. However, the integral over $c$ is intractable to compute. Therefore, an encoder approximation is introduced. For a training time step $i$, encoder distribution $q_{\phi } (c_{i}|x_{<i},t)$ is used to approximate the posterior distribution $p_{\theta  } (c_{i}|x_{\le i},t)$, where $\phi$ and $\theta$ is the parameters of the encoder (with attention model) and the autoregressive decoder. The KL-divergence of the encoder approximation from the posterior distribution can be written as Eq.\ref{4}. In Eq.\ref{5}, since the KL-divergence term is always non-negative, the conditional likelihood $\log_{}{p_{\theta  } (x_{i}|x_{<i},t)}$ equals the variational lower bound $\mathcal{L} (\theta,\phi,x,t)$ only when the KL-divergence $D_{KL} (q_{\phi } (c_{i}|x_{<i},t)\parallel p_{\theta} (c_{i}|x_{\le i},t))$ equals 0. This means that $c_{i}$ and $x_{i}$ are conditionally independent. As summarized in \cite{liu2019maximizing}, when $x_{<i}$ contains enough information to predict $x_{i}$, the model tends to reduce the dependence between $c_{i}$ and $x_{i}$ to maximize $\log_{}{p_{\theta  } (x_{i}|x_{<i},t)}$. At the same time, it also reduces the dependence between text and acoustic features. Modeling dependency between the text and the predicted acoustic features insufficiently may lead to higher bad-case rate. 

For the duration-based model, the dependence between time-aligned latent variables (generated by duration predictor with length regulator) and acoustic features is further reduced because there is no attention model to explicitly connect $c_{i}$ and $x_{<i}$. Introducing an auxiliary CTC recognizer to maximize the mutual information between the text and acoustic features is an optional way \cite{liu2019maximizing}. To alleviate these problems fundamentally, a non-autoregressive single-layer RNN is used as a decoder in this work. Due to the absence of autoregressive feedback, only hidden text representations are fed to the decoder which avoids the influence of local information preference, but also also puts forward a higher requirement for their representation ability. We believe that Dense-fuse encoder with strong text encoding ability can meet the requirement. In order to allow the text information to fully participate in the decoding process, the output of decoder and the expanded text representation are concatenated for final acoustic feature generation.



\vspace{-0.3cm}
\begin{equation}
\begin{aligned}
&\quad D_{KL} (q_{\phi } (c_{i}|x_{<i},t)\parallel p_{\theta} (c_{i}|x_{\le i},t))\\
&=\mathbb{E}_{q_{\phi }(c_{i}|x_{<i},t)}[\log_{}{q_{\phi} (c_{i}|x_{<i},t)}-\log_{}{p_{\theta  } (c_{i}|x_{\le i},t)} ]\\
&=\mathbb{E}_{q_{\phi } (c_{i}|x_{<i},t)}[\log_{}{q_{\phi } (c_{i}|x_{<i},t)}-\log_{}{p_{\theta  } (x_{i},c_{i}|x_{<i},t)}\\
&\quad +\log_{}{p_{\theta  } (x_{i}|x_{<i},t)}]\\
&=\mathbb{E}_{q_{\phi } (c_{i}|x_{<i},t)}[\log_{}{q_{\phi } (c_{i}|x_{<i},t)}-\log_{}{p_{\theta  } (x_{i},c_{i}|x_{<i},t)}]\\
&\quad +\log_{}{p_{\theta  } (x_{i}|x_{<i},t)}
\end{aligned}
\label{4}
\end{equation}
\vspace{-0.3cm}
\begin{equation}
\begin{aligned}
&\quad \log_{}{p_{\theta  } (x_{i}|x_{<i},t)}\\
&=D_{KL} (q_{\phi } (c_{i}|x_{<i},t)\parallel p_{\theta}(c_{i}|x_{\le i},t))\\
&\quad +\mathbb{E}_{q_{\phi }(c_{i}|x_{<i},t)}[\log_{}{p_{\theta  }(x_{i},c_{i}|x_{<i},t)}- \log_{}{q_{\phi }(c_{i}|x_{<i},t)}]\\
&\ge\mathbb{E}_{q_{\phi }(c_{i}|x_{<i},t)}[\log_{}{p_{\theta  }(x_{i},c_{i}|x_{<i},t)}- \log_{}{q_{\phi }(c_{i}|x_{<i},t)}]\\
&=\mathcal{L} (\theta,\phi,x,t)
\end{aligned}
\label{5}
\end{equation}
\section{Experiments}
\subsection{Datasets}
Experiments were conducted on two Mandarin corpora. The first one was recorded by a professional Chinese female speaker in studio quality. The transcription used in the recording covered multiple fields, with an average sentence length of 70 characters. The number of utterances used for training was 9600. The other one consisted of 12000 audio files extracted from an online audio-book which recorded by an actor with rich rhythm. The transcription was the corresponding novel, with an average sentence length of 86 characters. All recordings were sampled at 16kHz with 16-bit quantization. Consistent with LPCNet \cite{valin2019lpcnet}, 18 Bark cepstral coefficients and 2 pitch parameters were extracted as the prediction targets of Nana-HDR. 100 utterances that have not appeared in the training set were used for the in-domain naturalness testing. In order to verify whether the model can cope with out-domain scenarios, 200 popular words with an average length of 5 characters and 50 long paragraphs with an average length of 1000 characters were used for robustness testing. The latter was selected from WeChat official account, covering the fields of politics, sports, entertainment, literature, cooking and so on.
\subsection{Model configuration}
Nana-HDR was a sequence-to-sequence acoustic model with a Dense-fuse encoder, a duration predictor (with a length regulator) and a single-layer RNN-based decoder. The main components of the Dense-fuse encoder were: (1) A convolutional processor with three 256-dimensional Conv-1D layers whose kernel sizes were set to 3. (2) A text feature extractor with four Transformer blocks whose attention head numbers and hidden sizes were set to 4 and 256 respectively. (3) A 256-dimensional 4-head attention layer for fine feature fusion. Duration predictor was mainly composed of three 256-dimensional Conv-1D layers with kernel size of 3 and a 64-dimensional bidirectional GRU layer. The decoder was a non-autoregressive 512-dimensional unidirectional GRU layer. The postnet was a CBHG module with the same structure as the one in Tacotron. Nana-HDR was trained using the Adam optimiser \cite{kingma2015adam}. L1 loss was used for acoustic feature (before and after postnet) and duration loss. The model was trained for 300,000 steps, with a learning rate of 0.0001 and a batch size of 32. LPCNet as a relatively lightweight neural vocoder is applied in this work. Different from \cite{valin2019lpcnet}, in addition to being input to the first GRU layer, conditional feature was also fed to the second GRU layer.

\subsection{Results}
To evaluate the naturalness of the proposed model, we conducted Mean Opinion Score (MOS) and Comparison Mean Opinion Score (CMOS) tests. For all MOS tests, two groups of native Chinese speakers (5 in each group) were invited to listen and score 125 audio each time. 100 test utterances synthesized by the corresponding model were mixed with 25 original recordings, and the listener did not know which category each audio belonged to. Scores ranged from 1 to 5 with intervals of 0.5. The final MOS was obtained by averaging the scores of the two groups. For CMOS tests, the same listeners were asked to listen to the paired test utterances synthesized by two different models in random order and evaluate how the latter feels comparing to the former using a score in [-3, 3] with intervals of 1 (from much worse to much better). All listening tests were conducted in a quiet room with headphones. We evaluated the robustness by measuring the failure rate and the word error rate (WER). The failure was mainly identified by whether the synthesized audio ended early, repeated the same clip or contained meaningless clip which seriously affected the understanding of the content. The WER was measured by an ASR system described in \cite{wang2020transformer}. Relevant audio samples were available on the accompanying web page\footnote{https://linshilun.github.io/nanahdrsamples/nanahdr.html}. 

\subsubsection{Naturalness and robustness}
Our Nana-HDR was compared with two popular and well-performing models. Both models had the single type of encoder and decoder. Model I: Autoregressive attention-based Tacotron consisted of a CBHG encoder, a GMM-based attention model \cite{graves2013generating} which have been shown to improve the robustness \cite{battenberg2020location}, a two-layers autoregressive LSTM decoder and a CBHG postnet. Model II: Non-attentive non-autoregressive model Fastspeech had the same configuration as described in \cite{ren2019fastspeech}. It consisted of 6 FFT blocks on both the text side and the acoustic feature side and a duration predictor to connect the two parts. All models were well trained with the same number of iterations (300,000 steps) to ensure their performance.

Table~\ref{tab:mos} and Table~\ref{tab:cmos} contain MOS and CMOS results. It can be seen that Nana-HDR filled the naturalness gap between Fastspeech and Tacotron on both corpora. Listeners preferred the results synthesized by Nana-HDR to those synthesized by the other two systems. The results indicated that our Nana-HDR has achieved competitive performance in naturalness. In the aspect of robustness, because of the bad attention alignments, Tacotron with GMM-based attention had overall failure rates of 1.5\% and 2.8\% on two corpora. There was no serious synthesis failure in non-attentive models. General speech recognition was performed for synthesis samples without serious failure. WERs are recorded in Table~\ref{tab:wer}. The results indicate that Nana-HDR achieved the lowest WER whether trained with a studio quality corpus or with a challenging audio-book corpus.

We attributed three possible reasons why Nana-HDR can achieve good naturalness and robustness. First, the hybrid structure without attention model; secondly, the strong text feature extraction ability of the Dense-fuse encoder; thirdly, stable conditional generation ability of non-autoregressive stateful RNN decoder. The previous experimental results demonstrated the advantages of non-attentive structure with hybrid Transformer-based encoder and RNN-based decoder. Next, we conducted ablation studies to verify the effectiveness of Dense-fuse encoder and non-autoregressive RNN-based decoder in Nana-HDR. The results are shown in the table~\ref{tab:amos},~\ref{tab:acmos} and ~\ref{tab:awer}.
\vspace{-0.5cm}
\begin{table}[!htb]
  \caption{MOS with 95\% confidence intervals.}
  \label{tab:mos}
  \centering
  \setlength{\abovecaptionskip}{0.1cm}
  \begin{tabular}{ccc}
    \toprule
    Model & female & male \\
    \midrule
    Fastspeech & $ 4.10\pm0.05$ & $ 4.08\pm0.06 $ \\
    Tacotron-GMMA & $ 4.20\pm0.04$ & $ 4.13\pm0.06 $ \\
    Nana-HDR & $ \textbf{4.22}\pm\textbf{0.04}$ & $ \textbf{4.23}\pm\textbf{0.05}$ \\
    Ground truth & $ 4.41\pm0.04$ & $ 4.37\pm0.04 $ \\
    \bottomrule
  \end{tabular}
\end{table}
\vspace{-0.7cm}
\begin{table}[!htb]
  \caption{CMOS comparison, the p-value of t-test between the two systems p\textless0.01.}
  \label{tab:cmos}
  \centering
  \setlength{\abovecaptionskip}{0.1cm}
  \begin{tabular}{ccc}
    \toprule
    Model & female & male \\
    \midrule
    Nana-HDR & \textbf{0.000}  & \textbf{0.000} \\
    Tacotron-GMMA & -0.245 & -0.311 \\
    Fastspeech & -0.336  & -0.363 \\
    \bottomrule
  \end{tabular}
\end{table}
\vspace{-0.7cm}
\begin{table}[!htb]
  \caption{Word error rate of the neural TTS models.}
  \label{tab:wer}
  \centering
  \setlength{\abovecaptionskip}{0.1cm}
  \begin{tabular}{ccc}
    \toprule
    Model & female & male \\
    \midrule
    Tacotron-GMMA & 2.8\% & 4.1\% \\
    Fastspeech & 2.7\%  & 3.4\% \\
    Nana-HDR &  \textbf{2.0\%}  & \textbf{2.1\%} \\
    \bottomrule
  \end{tabular}
\end{table}
\vspace{-0.6cm}
\subsubsection{Ablation studies}
First, it was found that the duration extractor often failed to align when feature fusions were removed from the Dense-fuse encoder. And the absence of fine feature fusion led to inaccurate duration extraction of long pause. Therefore, we used the duration extractor with the original Dense-fuse encoder in subsequent experiments. Removing feature fusions from Nana-HDR (Nana-HDR-Nofusion) resulted in -0.153 and -0.258 CMOS. The WERs on both corpora were also higher. Then, we replaced the Dense-fuse encoder with the commonly used RNN-based CBHG encoder (Nana-HDR-CBHGE). This reduced the MOS on two corpora by 1.7\% and 5.2\% respectively. In terms of CMOS, listeners preferred the results synthesized by the system without replacement. Using CBHG encoder also resulted in higher WERs on both corpora. We argued that the output of the Dense-fuse encoder obtained better representation ability which made Nana-HDR perform better. On the one hand, rich linguistic information helped to improve the naturalness. On the other hand, it was more suitable for the text encoding with stronger representation capability to be the only input of non-autoregressive RNN-based decoder.

Then we replaced the non-autoregressive RNN-based decoder with an autoregressive one (Nana-HDR-ARD) which led to significant degradation in naturalness, robustness and speed (reduced by 8.85x). On the challenging audio-book corpus, we found that there were obvious word skipping and mispronunciation in some in-domain sentences, which has not been seen in other experiments. We held that autoregressive decoder without frame reduction was more sensitive to the exposure bias and the local information preference, which increased the bad case rate.

Ablation studies demonstrated that Dense-fuse encoder and non-autoregressive RNN-based decoder were effective components of Nana-HDR. It also showed that when the text information was fully extracted, it was feasible to use an efficient non-autoregressive RNN decoder to eliminate the influence of exposure bias and local information preference fundamentally.
\vspace{-0.4cm}
\begin{table}[!htb]
  \caption{MOS in the ablation studies with 95\% confidence intervals.}
  \label{tab:amos}
  \centering
  \setlength{\abovecaptionskip}{0.1cm}
  \begin{tabular}{ccc}
    \toprule
    Model & female & male \\
    \midrule
    Nana-HDR-Nofusion & $ 4.20\pm0.04$ & $ 4.18\pm0.05$ \\
    Nana-HDR-CBHGE & $ 4.15\pm0.06$ & $ 4.01\pm0.06 $ \\
    Nana-HDR-ARD & $ 4.11\pm0.08$ & $ 3.73\pm0.07 $ \\
    \bottomrule
  \end{tabular}
\end{table}
\vspace{-0.4cm}
\begin{table}[!htb]
  \caption{CMOS comparison in the ablation studies, the p-value of t-test between the two systems p\textless0.01.}
  \label{tab:acmos}
  \centering
  \setlength{\abovecaptionskip}{0.1cm}
  \begin{tabular}{ccc}
    \toprule
    Model & female & male \\
    \midrule
    Nana-HDR-Nofusion & -0.153  & -0.258 \\
    Nana-HDR-CBHGE & -0.265  & -0.471 \\
    Nana-HDR-ARD & -0.349 & -0.694 \\
    \bottomrule
  \end{tabular}
\end{table}
\vspace{-0.4cm}
\begin{table}[!htb]
  \caption{Word error rate of models in the ablation studies.}
  \label{tab:awer}
  \centering
  \setlength{\abovecaptionskip}{0.1cm}
  \begin{tabular}{ccc}
    \toprule
    Model & female & male \\
    \midrule
    Nana-HDR-Nofusion & 2.2\%  & 2.4\% \\
    Nana-HDR-CBHGE & 2.5\%  & 3.3\% \\
    Nana-HDR-ARD & 2.5\% & 4.4\% \\
    \bottomrule
  \end{tabular}
\end{table}
\vspace{-0.4cm}

\section{Conclusions}
In this paper, we proposed Nana-HDR, a non-attentive non-autoregressive hybrid model for TTS. It improves the representation ability of text encoding by a Transformer-based Dense-fuse encoder. The influence of the exposure bias and the local information preference can be avoided through the non-autoregressive RNN-based decoder. Collapse caused by attention failure can be avoided by the non-attentive structure. By fully exploiting the advantages of each component, Nana-HDR achieves competitive performance in both naturalness and robustness compared with Tacotron and Fastspeech.

\bibliographystyle{IEEEbib}
\bibliography{strings,refs}

\end{document}